\documentclass{article}
\usepackage{microtype}
\usepackage{graphicx}
\usepackage{subcaption}
\usepackage{booktabs} 
\usepackage{tikz}
\usepackage{xcolor}
\usepackage{amsmath}
\usepackage{multirow}
\usepackage{ragged2e}
\usepackage{pgfplots}
\pgfplotsset{compat=1.18}
\usepackage{listings} 
\usetikzlibrary{positioning, shapes.geometric, arrows.meta}
\usepackage{fontawesome5}
\usepackage{caption}
\usepackage{natbib}
\usepackage{hyperref}

\usepackage{tcolorbox} 
\usepackage{xurl}
\usepackage{longtable} 
\usepackage{textcomp}  
\tcbuselibrary{breakable}

\usepackage[accepted]{icml2025}
\usepackage{amssymb}
\usepackage{mathtools}
\usepackage{amsthm}
\usepackage[capitalize,noabbrev]{cleveref}
\usepackage{enumitem}
\usepackage{anyfontsize}
\usepackage{multicol}
\theoremstyle{plain}

\usepackage{booktabs}
\usepackage{longtable}
\usepackage{array}
\usepackage{siunitx}

\theoremstyle{definition}

\theoremstyle{remark}

\usepackage[textsize=tiny]{todonotes}
\icmltitlerunning{\textsc{InvestLogicBench2026}}
\begin{document}
\twocolumn[
\icmltitle{Evaluating Investment Logic in Large Language Models: A Real-World Benchmark Towards Personalzied Financial Agents}
\icmlsetsymbol{equal}{*}
\icmlsetsymbol{dagger}{$\dagger$}

\begin{icmlauthorlist}
\icmlauthor{Yuanhong Jiang}{equal,dagger,sch2,HiThink}
\icmlauthor{Jingjie Zou}{equal,DIG Group}
\icmlauthor{Rui Jiang}{HiThink}
\icmlauthor{Zhenghong Lin}{sch2}
\icmlauthor{Xusheng Yu}{sch}
\icmlauthor{Qiqi Huang}{sch}
\icmlauthor{Shuai Jia}{HiThink}
\icmlauthor{Shijie Dai}{dagger,tongji,HiThink}
\end{icmlauthorlist}

\icmlaffiliation{tongji}{Tongji University, Shanghai, China}
\icmlaffiliation{HiThink}{HiThink Research}
\icmlaffiliation{DIG Group}{DIG Group, Guangdong, China}
\icmlaffiliation{sch}{ Fuzhou University, Fuzhou, China}
\icmlaffiliation{sch2}{Zhejiang University, Hangzhou, China}
\icmlcorrespondingauthor{Yuanhong Jiang}{jiangyuanhong@myhexin.com}
\icmlcorrespondingauthor{Shijie Dai}{daishijie@myhexin.com}

\icmlkeywords{Investment Logic,Market-aligned Agent,KOL Investors,SIL Engine}
\vskip 0.3in
]


\begin{abstract}
The dominant paradigm for evaluating Large Language Models remains anchored in chat and question-answering (QA) tasks, creating a fundamental mismatch with the competencies required for reliable agent execution in real-world domains. This disconnect is acutely evident in financial decision-making, where proficiency is defined not by knowledge recall, but by the capacity to align \textbf{Person}—the investor's unique profile, constraints, and risk preferences—with the interpretation of unstructured market \textbf{Events} (E), the synthesis of coherent investment \textbf{Reasoning} (R), the formulation of corresponding \textbf{Decisions} (D), and the pursuit of profitable \textbf{Outcomes} (O). Without evaluation grounded in this integrated P$\rightarrow$E$\rightarrow$R$\rightarrow$D$\rightarrow$O
chain, progress toward capable financial agents remains stalled.

To bridge this gap, we introduce \textsc{InvestLogicBench2026}, a benchmark derived from the documented decision-making of real-world investors. It provides structured, validated traces that explicitly instantiate such investment logic chain for each investment thesis. Our evaluation reveals a pronounced investment logic gap: state-of-the-art LLMs encounter significant bottlenecks in both emulating the reasoning of expert investors and translating such reasoning into profitable outcomes. Our benchmark establishes the first reproducible and market-aligned standard for quantifying this deficiency. This work shifts the evaluation paradigm from passive knowledge recall to active reasoning, paving the way for building robust, end-to-end LLM investment agents that generate market-aligned and personalized decisions adaptable to diverse strategies. The project is publicly accessible in \href{url}{https://anonymous.4open.science/r/InvestLogicBench-C2D7}.
\end{abstract}

\section{Introduction}
\label{sec:intro}

The rapid integration of Large Language Models (LLMs) into finance has transformed tasks ranging from sentiment analysis and earnings call summarization to macroeconomic forecasting, yet their deployment in live trading environments reveals significant performance limitations. This progression has culminated in the emergence of live-trading arenas where LLM-based agents compete with real capital, relying solely on real-time news and market data \citep{fan2025aitrader}. These platforms provide an unforgiving, forward-looking test of practical capability, revealing a striking dissonance: models that excel on established financial reasoning benchmarks often generate unprofitable or unstable strategies in live markets.

The observed ``Capabilities-Performance Paradox''\citep{ding2024survey} points to a fundamental distinction between financial knowledge and investment intelligence. Current LLMs demonstrate impressive proficiency at retrieving facts, solving textbook problems, and performing quantitative calculations---skills meticulously measured by benchmarks such as FinBen \citep{xie2024finben} and BizBench \citep{koncelkedziorski2024bizbench}. Yet they lack the higher-order logic required for successful market participation: the ability to distill salient signals from unstructured news, connect disparate geopolitical, corporate, and macroeconomic events into a coherent narrative, adjudicate between conflicting data points under uncertainty, and formulate---then steadfastly maintain---a strategic thesis that aligns with a consistent investment style.

We argue that this failure is not incidental but systemic, stemming from three \textbf{structural mismatches} between the prevailing LLM paradigm and the cognitive requirements of successful investing:

\textbf{Objective Mismatch:} LLMs are trained via next-token prediction on vast corpora, optimizing for linguistic coherence over causal reasoning. Raw financial text, while abundant, rarely contains explicit annotations of the investment decision cycle: how an investor's unique \textbf{Person} (P)—comprising their objectives, risk constraints, and behavioral biases—interacts with market \textbf{Events} (E), which are then interpreted and synthesized into investment \textbf{Reasoning} (R), leading to corresponding \textbf{Decisions} (D), and ultimately aiming at profitable \textbf{Outcomes} (O). Without exposure to structured P$\rightarrow$E$\rightarrow$R$\rightarrow$D$\rightarrow$O investment logic chains, models struggle to internalize the instrumental logic of investment decision-making.
 \citep{chen2024distilling}.

\textbf{Evaluation Gap:} Existing financial benchmarks overwhelmingly assess static QA skills---fact retrieval, calculation, and well-defined problem-solving. They evaluate a model's prowess as a financial analyst but fail to assess its competency as a portfolio manager: its ability to identify consequential signals in real-time noise, synthesize narratives from unfolding events, and maintain strategic discipline. This creates an \textbf{evaluation opacity}: we can measure whether an agent succeeds (via P\&L) but cannot diagnose \textbf{the underlying reasoning quality}, thus obscuring whether success stems from sound logic or stochastic luck \citep{ding2024survey}.

To bridge this chasm, we shift the paradigm from evaluating what models know to assessing how they think. We introduce InvestLogicBench2026, the first benchmark constructed from the dynamic, articulated reasoning processes of successful market participants. It provides structured, validated traces that explicitly instantiate the P$\rightarrow$E$\rightarrow$R$\rightarrow$D$\rightarrow$O chain for each investment thesis.
Our framework extracts and structures investment theses from experienced fund managers and key opinion leaders (KOLs) into explicit P$\rightarrow$E$\rightarrow$R$\rightarrow$D$\rightarrow$O chains, capturing market events, logical rationale, concrete actions, and market-verified outcomes.

Our contributions are threefold:
\begin{enumerate}
    \item Through a seven-week experimental study comparing live agent profitability with benchmark results, we empirically identify a pronounced capability-performance paradox in leading LLMs when applied to real-world market investment.
    \item We present \textsc{InvestLogicBench2026}, the first benchmark for evaluating practical investment logic, featuring 201,247 structured decisions annotated with events, reasoning , decisions, actions, and market-verified outcomes from 151 real-world investors.
    \item We quantify a significant investment logic gap: leading LLMs achieve only 1.8/5 reasoning score against expert traces, revealing systematic failures in event-noise discrimination and long-horizon consistency that explain their live trading underperformance.
\end{enumerate}

By establishing a diagnostic benchmark for market-aligned reasoning under uncertainty, \textsc{InvestLogicBench2026} shifts the evaluation focus from financial knowledge recall to the critical ability to formulate and execute coherent investment theses. 

\section{Can Language Models Make Profitable Investment in Real-World Markets?}

\begin{figure*}[t]
\centering
\includegraphics[width=1\textwidth]{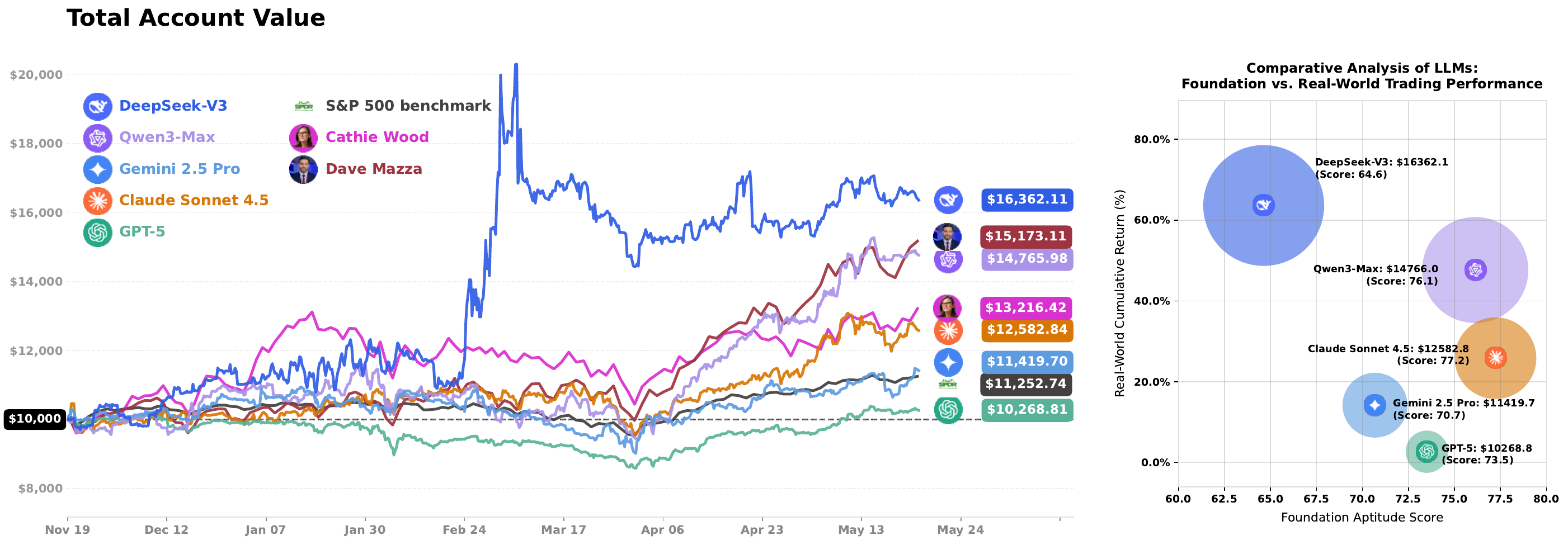}
\vspace{-0.7cm}
\caption{\textbf{The ``Capabilities-Performance Paradox'' in financial domain.} (Left) Cumulative equity curves between autonomous LLM agents and human expert proxies in a 7-week live U.S. stock trading competition. DeepSeek-V3 achieved the highest cumulative return, significantly outperforming the S\&P 500 benchmark, while models like GPT-5 and Claude-Sonnet-4.5 demonstrated lackluster performance in seizing opportunities. Key market inflection points highlight periods where expert logic diverged from model behavior. (Right) Comparative analysis between Foundation Aptitude Scores and Real-World Cumulative Returns. It suggests that superior performance on general benchmarks fails to reliably transfer into financial analytical acumen or effective real-world trading execution.}
\label{fig:live_vs_benchmark}
\end{figure*}

Despite demonstrating outstanding competence in information retrieval and accessing specialist domain knowledge, current state-of-the-art models frequently struggle to capture critical financial events and conduct rigorous logic analysis when navigating the stochastic, adversarial, and non-stationary dynamics of live trading scenarios, which we termed the ``Capabilities-Performance Paradox''. This disconnect highlights a fundamental reliability gap: the proficiency to explain financial concepts does not guarantee the agency required to analyze and execute profitable strategies under uncertainty.

\subsection{The U. S. Equity Trading Arena: A Live-Fire Market Competition}
To rigorously evaluate the investment logic of LLMs in market-aligned conditions, we instantiated a live U.S. equity trading arena. From November 19, 2025, multiple LLM-based agents, each initialized with \$10,000 in simulated capital, competed autonomously to maximize risk-adjusted returns through discretionary hourly trading of U.S. stocks and ETFs.

The arena was designed to test the complete P$\rightarrow$E$\rightarrow$R$\rightarrow$D$\rightarrow$O chain under uncertainty. Agents maintained strategic continuity through access to their full \texttt{Portfolio Report} and \texttt{Trading History}. To simulate professional practice, they actively foraged for information via a dual-tool \textbf{Perception-Action Loop}—querying structured data with \texttt{FinQuery} and unstructured intelligence with \texttt{Search}—then synthesized these signals into executable trades with explicit reasoning. This forward-testing format eliminated retrospective bias, providing a pure assessment of an LLM's ability to form and act on market logic in real time.

\subsection{The Capabilities-Performance Paradox}
As visualized in Figure \ref{fig:live_vs_benchmark} (Right), we contrasted the models' Foundation Aptitude Scores—an average score derived from MMLU-Pro, SWE-bench, AIME 2025, and SimpleQA—against their realized financial returns. While scaling laws typically postulate a positive correlation between general reasoning capabilities and downstream task performance, our empirical data exhibits a distinct counterintuitive misalignment.

Strikingly, models possessing superior foundation scores, such as Claude-Sonnet-4.5 (Score: 77.2) and GPT-5 (Score: 73.5), failed to translate their general reasoning dominance into effective market execution. Throughout the \textbf{7-week live trading window (Nov. 19, 2025 -- Jan. 9, 2026)}, GPT-5 ended with a portfolio value of \$9,901.1 (-1.0\%), while Claude-Sonnet-4.5 yielded negligible gains, closing at \$10,072.8 (+0.7\%). Both models failed to surpass the passive S\&P 500 (SPY) benchmark, which returned approximately +4.3\% over the same period, indicating that most models were paradoxically less effective in the live financial domain. In sharp contrast, as detailed in the cumulative equity curves in Figure \ref{fig:live_vs_benchmark} (Left), models with comparatively lower general evaluation scores demonstrated significantly stronger market adaptability. DeepSeek-V3 achieved the highest cumulative return of +14.2\% (\$11,420.3), significantly outperforming both the market baseline and its higher-scoring counterparts.

This divergence suggests a critical transfer barrier, where current generalist metrics fail to capture the nuances of financial intelligence-specifically the ability to synthesize noisy, non-stationary signals into profitable decisions. This finding exposes a fundamental gap in current evaluation paradigms: the proficiency to solve static problems is decoupled from the agency required to execute rigorous financial logic under uncertainty.

\subsection{Deconstructing the Divergence: Consensus Bias and Reasoning Fragmentation}
To diagnose the underlying mechanisms of the ``Capabilities-Performance Paradox'', we conduct a retrospective analysis of the trading period and identify four pivotal market events that exerted the key influence on U.S. market.
\begin{itemize}
    \item \textbf{Hawkish Minutes \& Nvidia Volatility (Nov. 19--20, 2025): } Hawkish FOMC minutes dampened rate cut hopes, coinciding with a reversal in Nvidia stock. The superposition of fading liquidity support and reignited AI valuation fears triggered a sharp sell-off.
    \item \textbf{Fed Delivers Hawkish Cut (Dec. 9, 2025):} The Fed delivered a widely expected 25 basis point cut with only one cut for 2026. The lack of a fresh positive catalyst triggered a ``sell-the-news'' dynamic with subsequent market adjustment.
    \item \textbf{CPI Surprise Cools Anxiety (Dec. 18, 2025):} November CPI cooled to 2.7\% (3\% forecast), reigniting expectations for further rate cuts. This data thoroughly eliminated market tension, acting as the catalyst for a sustained subsequent rally.
    \item \textbf{Post-Correction Rebound (Jan. 2, 2026): } Year-end profit-taking and selling pressure were fully digested, forming a consensus around a soft landing. This renewed optimism launched a fresh rally to start the year.
\end{itemize}

By decomposing the aforementioned turning points and the timeline, we observed that while the broader market (SPY) reacted dynamically to these pivotal events, models like GPT-5 and Claude-Sonnet-4.5 failed to correctly interpret the signal, resulting in flatlining performance.

Initially, the homogenous drawdowns observed during the ``Hawkish Minutes \& Nvidia Volatility'' phase suggest a shared baseline risk sensitivity across all agents. However, a significant bifurcation emerged during the subsequent market rally. Claude-Sonnet-4.5 and GPT-5 executed a premature risk-off rotation—effectively liquidating positions during the transient shock—thereby failing to capture the ensuing liquidity-driven upside.

Specific inefficiencies were observed in GPT-5, which incurred a maximum drawdown exceeding the SPY benchmark following the ``Fed Hawkish Cut''. Crucially, it failed to re-enter the market effectively during the ``CPI Surprise'' and ``Post-Correction Rebound'' phases. In contrast, despite Qwen3-Max suffering a comparable drawdown, it exhibited a rapid recovery dynamic, repairing losses immediately as market sentiment shifted. Notably, DeepSeek-V3 displayed advanced strategic timing effectively accumulating assets during the market cooling phase and realizing outsized returns during the recovery.

By comparing the decision traces of LLMs, two fundamental cognitive failures—Consensus Bias and Reasoning Fragmentation—are revealed as the drivers of the capabilities-performance paradox. Models like GPT-5 and Claude-Sonnet-4.5 tend to converge toward the statistical mean of their training data to maximize token likelihood; this results in merely reproducing priced-in market consensus rather than generating the variant perception necessary for alpha. Simultaneously, when confronted with a complex, non-stationary environment, these agents exhibited Reasoning Fragmentation. Overwhelmed by the volume of exogenous data, they reverted to rigid, myopic thinking patterns that failed to account for forward-looking dynamics, ultimately missing the pivotal rebound that drove the benchmark's outperformance.

\subsection{The Human Benchmark: Establishing Expert Logic Anchors}
The divergent performance of LLMs raises a critical question: How would human experts perform during this trading period? To incorporate expert logic, we introduced two professional investors:
\vspace{-0.3cm}
\begin{itemize}
    \item \textbf{Bill Miller:} The founder of Miller Value Partners, renowned for his legendary 15-year streak of beating the S\&P 500 (Emulate through the Miller Value Partners Appreciation ETF).
    \item \textbf{Cathie Wood:} The founder and CEO of ARK Invest, who commits to find technologies where declining costs trigger exponential demand growth (Emulate through the ARK Innovation ETF).
\end{itemize}
\vspace{-0.1cm}

\begin{figure*}[t]
    \centering
    \includegraphics[width=0.95\textwidth]{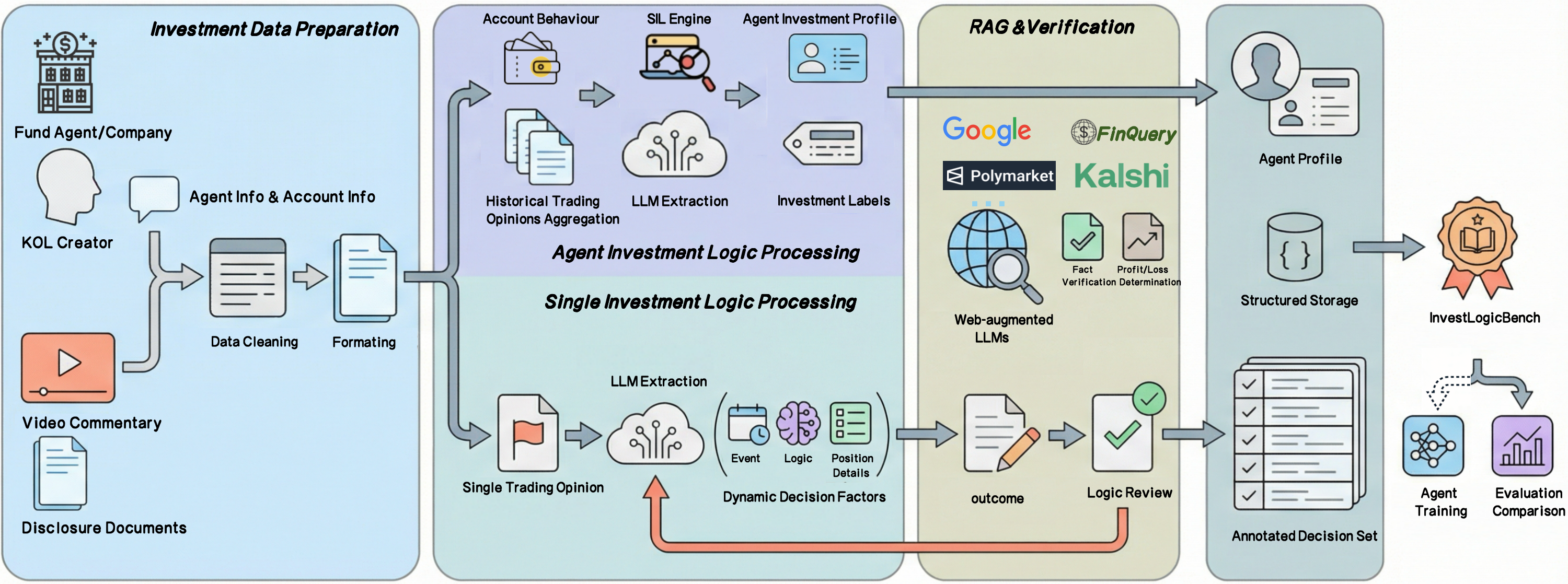}
    \vspace{-0.3cm}
    \caption{The two-stage construction pipeline of \textsc{InvestLogicBench2026}. Stage 1 creates a static investor profile by aggregating multiple data points from a single agent. Stage 2 performs per-decision annotation, which includes a quality control loop where logic is verified and potentially fed back to refine the profile.}
    \label{fig:pipeline}
\end{figure*}

As illustrated in Figure \ref{fig:live_vs_benchmark} (Left), human expert portfolios demonstrated superior resilience and excess return generation. During critical shocks like the ``Hawkish Cut'' and the ``Nvidia Volatility'', the experts exhibited counter-cyclical aggression. Unlike most investors and consensus-following LLMs, human experts can employ forward-looking reasoning to avoid drawdowns and accelerate decisively when opportunities arise. This allowed them to capture the full magnitude of the ``Post-Correction Rebound'' with pleasing cumulative returns (e.g., Miller at +10\% and Wood at +8\%).

The deconstruction of experts' unique decision-making chains provides critical evidence for understanding the performance deficiencies of LLMs in dynamic gaming scenarios. Deep analysis of this divergence reveals that the expert advantage stems from variant perception. When the Key Events occurred, the experts leveraged an independent reasoning framework to gather information and identify the mispricing, whereas LLMs lack this depth of analysis. Integrating human experts' analysis logic is essential not just for comparison, but as a diagnostic reference system. 

\section{The \textsc{InvestLogicBench2026} Dataset}
\label{sec:dataset}

\subsection{Design and Construction}

\textsc{InvestLogicBench2026} is constructed from the dynamic decision-making processes of influential real-world investment agents (e.g., major KOLs and fund managers), whose viewpoints are widely followed and accepted by substantial investor communities. Its design is built upon the \textbf{E→R→D→O investment logic chain}, where market \textit{Events} (E) inform \textit{Reasoning} (R) leading to a \textit{Decision} (D) aimed at an profitable \textit{Outcome} (O). This framework shifts evaluation from static question-answering to the assessment of active reasoning required for real-world investing.
\begin{figure*}[t]
    \centering
    \includegraphics[width=0.9\textwidth]{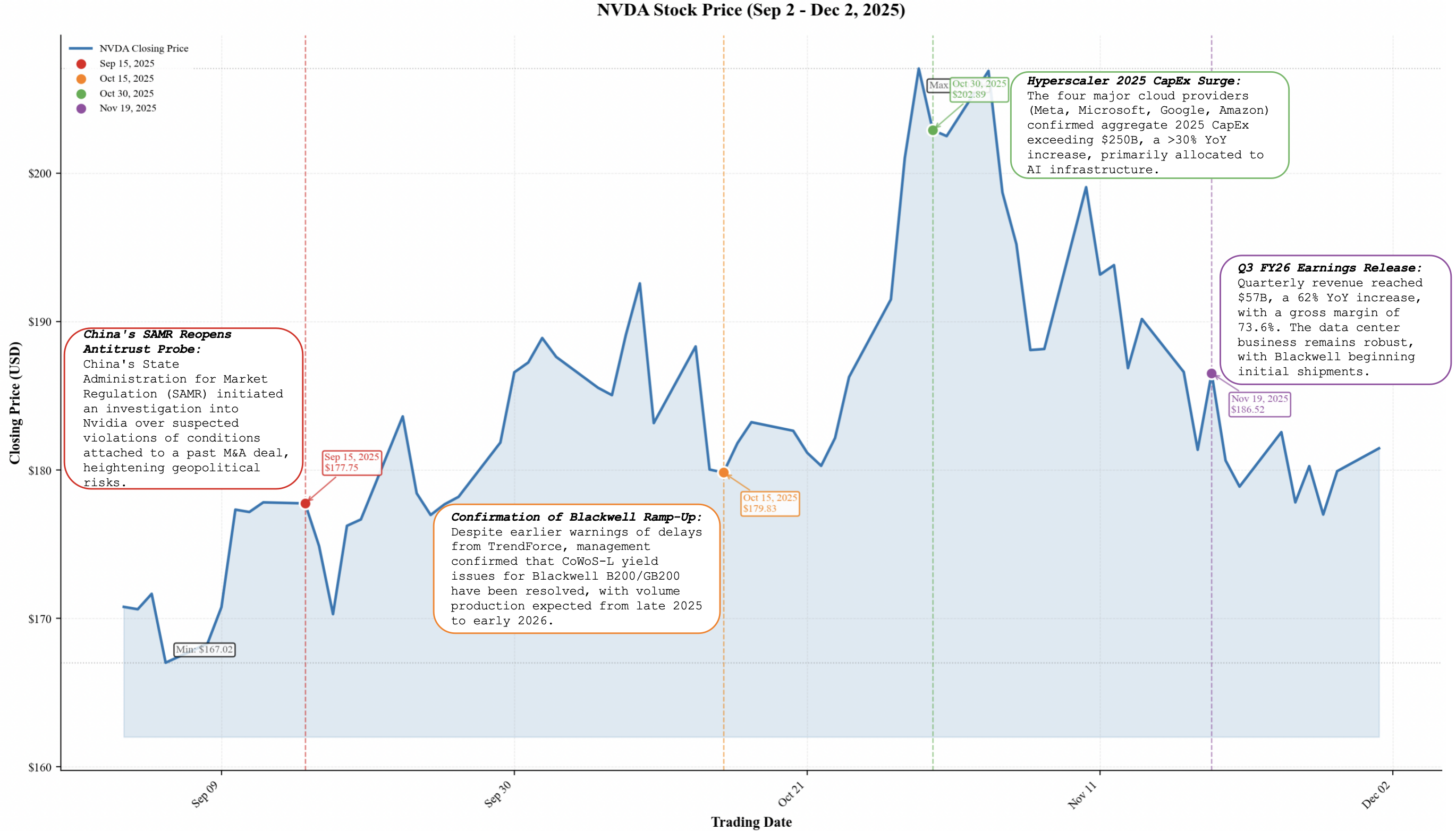}
    \vspace{-0.4cm}
    \caption{A working example from the Stock Investment Logic (SIL) Engine. It demonstrates how a specific market event (here, an antitrust probe announcement by China's SAMR targeting Nvidia on \texttt{2025-09-15}) is temporally linked to a security (\texttt{NVDA}) and its subsequent price series. The engine structures and stores such Event-Security-Time tuples to form a queryable knowledge base for inferring recurring investment logic patterns and enriching agent profiles.}
    \label{fig:sil_engine_example}
\end{figure*}

The benchmark is built through a four-stage pipeline (Fig.~\ref{fig:pipeline}): \textbf{1) Data Preparation} ingests multi-source data (e.g., video commentary, fund reports). \textbf{2) Agent Logic Processing} employs a dual-channel analysis: (i) extracting explicit investment logic from the agent’s own verbalized theses and (ii) using our proprietary \textbf{Stock Investment Logic (SIL) Engine} to infer implicit logic patterns from an agent's trading history, constructing a holistic behavioral profile. \textbf{3) RAG-Augmented Verification} ensures fidelity by checking logic based on real-market data including general search engine, financial database, and event prediction markets where event contracts are traded (e.g., Polymarket, Kalshi). \textbf{4) Benchmark Assembly} integrates verified decisions with agent profiles into a structured, queryable knowledge graph.

\subsection{Structured Schema}

Each data instance is a JSON object containing an \texttt{agentProfile} and a list of \texttt{decisions}, instantiating the E→R→D→O chain.

\noindent\textbf{Profile Fields} characterize the agent: \texttt{agentProfile} (personalized investment style and preference), \texttt{capital} (scale), \texttt{security} (asset classes), and metadata (\texttt{name}, \texttt{link}).

\noindent\textbf{Decision Fields} capture each logic-action triplet:
\begin{itemize}[noitemsep, topsep=2pt]
    \item \texttt{date}, \texttt{events}: Decision timestamp and catalyst events.
    \item \texttt{finLogic}: The complete E→R→D chain (e.g., '[Event] → [Thesis] → Action: [Action]').
    \item \texttt{investDecision}: The executable action (e.g., 'Long NVDA').
    \item \texttt{investPeriod}, \texttt{endDate}: Evaluation horizon and validation date.
    \item \texttt{outcome}: The market-validated outcome (e.g., '+5.2\%').
    \item \texttt{finLogicReview}: System-generated post-mortem analysis.
\end{itemize}

\subsection{Statistics and Tasks}

The inaugural version of \textsc{InvestLogicBench2026} comprises 201,247 annotated investment theses sourced from 151 distinct financial KOLs (data details in Appendix \ref{app:sample} and \ref{app:kol_urls}). The statistical summary of the investor profiles, which underscores the dataset's focus on real-world, event-driven strategies, is presented in Table \ref{tab:dataset_stats}.

\subsection{Benchmark Tasks}
We define three progressive tasks:
1.  \textbf{Logic Comprehension}: Given agent text, extract and summarize the \texttt{finLogic}. (Evaluates understanding)
2.  \textbf{Logic \& Thesis Generation}: Given current \texttt{events} and an \texttt{agentProfile}, generate a plausible \texttt{finLogic} and decision. (Evaluates synthetic reasoning)
3.  \textbf{End-to-End Investment Simulation}: According to the investment decisions, the outcome is evaluated with real-world market data. Both static and dynamic investment actions are evaluated depending on whether the account's position (holdings) status is disclosed. 

\subsection{Automated Evaluation via Judge LLM}
A high-capacity LLM (e.g., GPT-5) serves as an automated judge, scoring outputs based on:
- \textbf{Fidelity}: Similarity to ground-truth agent logic (Task 1).
- \textbf{Coherence \& Plausibility}: Logical soundness and market relevance (Tasks 2, 3).
- \textbf{Consistency}: Adherence to initial style and logical through-line (Task 3).
\begin{table}[htbp]
\centering
\small
\caption{Statistical Summary of Agent Profiles in \textsc{InvestLogicBench2026}}
\label{tab:dataset_stats}
\begin{tabular}{l|c}
\toprule
\textbf{Category} & \textbf{Statistics} \\
\midrule
Total Investment Experts & 151 \\
Total Annotated Decisions & 201,247 \\
Avg. Decisions per KOL & 1332.8 \\
Avg. Decisions per fund agent & 2.5 \\
\midrule
\textbf{Primary Style (\texttt{agentProfile})} & \\
\quad - Event-Driven & 102 (67.5\%) \\
\quad - Long-Term Value & 16 (10.6\%) \\
\quad - Technical & 8 (\textbf{5.3\%}) \\
\quad - Macro & 30 (\textbf{19.9\%}) \\
\midrule
\textbf{Capital Scale (\texttt{capital})} & \\
\quad - Below \$1M & 19 (\textbf{12.6\%}) \\
\quad - \$1M - \$10M & 41 (\textbf{27.2\%}) \\
\quad - \$10M - \$100M & 74 (\textbf{49.0\%}) \\
\quad - Above \$100M & 17 (\textbf{11.3\%}) \\
\midrule
\textbf{Primary Security Type (\texttt{security})} & \\
\quad - Equity & 42 (27.8\%) \\
\quad - Options & 16 (\textbf{10.6\%}) \\
\quad - Futures/Commodities & 6 (\textbf{4.0\%}) \\
\quad - ETF & 15 (9.9\%) \\
\quad - \textit{Others/Combined} & \textit{72 (47.7\%)} \\
\midrule
\textbf{Primary Sample Type} & \\
\quad - Event-driven decisions & 94.0\% \\
\quad - Forward-looking event predictions & 3.0\% \\
\quad \qquad Web-verifiable & 2.9\% \\
\quad \qquad Prediction-market verified & 0.1\% \\
\quad - Technical/quantitative analysis & 3.0\% \\
\bottomrule
\end{tabular}
\end{table}

\section{Empirical Analysis: The Investment Logic Gap}

We evaluate some leading LLMs on \textsc{InvestLogicBench2026} using Task 2 (Logic \& Thesis Generation), where models must synthesize an investment thesis from retrieved market events and an agent profile. Performance is measured along two axes: (1) \textbf{Reasoning Quality}—comprising \textit{Event Coherence} (fidelity to input events) and \textit{Logical Plausibility} (internal validity of the E$\to$R$\to$D chain), and (2) \textbf{Decision Outcome}—comprising \textit{Win Rate} and simulated \textit{Total Profit} over the investment horizon.

\subsubsection{Conditional Scoring Framework}
The overall score synthesizes reasoning quality and decision outcome. The \textbf{Decision Outcome Score} is a weighted composite of two critical performance dimensions: the agent's absolute profitability and its strategy win rate, each evaluated relative to the human expert benchmark.

First, the \textit{Profit Score} (\(S_{\text{profit}}\)) is calculated conditionally based on the agent's total profit (\(P_a\)) against the human expert benchmark (\(P_h\)) and a neutral midpoint (\(P_m = \max(0, P_h)\)):
\[
S_{\text{profit}} =
\begin{cases}
5.0, & \text{if } P_a \geq P_h \\
5.0 \times \dfrac{P_a - P_m}{P_h - P_m}, & \text{if } P_m \leq P_a < P_h \\
5.0 \times \dfrac{P_a}{P_m}, & \text{if } 0 \leq P_a < P_m \\
0.0, & \text{if } P_a < 0
\end{cases}
\]

Second, the \textit{Win Rate Score} (\(S_{\text{win}}\)) measures the consistency of positive returns. The agent's win rate (\(W_a\)) is compared to the human's (\(W_h\)):
\[
S_{\text{win}} =
\begin{cases}
5.0, & \text{if } W_a \geq W_h \\
5.0 \times \dfrac{W_a}{W_h}, & \text{if } W_a < W_h
\end{cases}
\]

The final \textbf{Decision Outcome Score} is the weighted sum of these components, emphasizing absolute profit:
\[
S_{\text{outcome}} = 0.7 \times S_{\text{profit}} + 0.3 \times S_{\text{win}}
\]

The \textbf{Reasoning Quality Score} remains the average of its \textit{Event Coherence} (EC) and \textit{Logical Plausibility} (LP), each on a 5-point scale: \(S_{\text{reasoning}} = (S_{\text{EC}} + S_{\text{LP}}) / 2\).

The final \textbf{Overall Score} is the sum of the reasoning and outcome scores, providing a balanced assessment on a 0-10 scale:
\[
S_{\text{overall}} = \underbrace{S_{\text{reasoning}}}_{\text{(0–5)}} + \underbrace{S_{\text{outcome}}}_{\text{(0–5)}}
\]
This composite metric ensures a robust and multidimensional evaluation of an LLM's investment logic, equally valuing the quality of its reasoning process and the practical efficacy of its decisions.

\subsubsection{Evaluation Results}
As shown in Table~\ref{tab:comprehensive_results}, four of the five evaluated models exceeded the expert profitability benchmark, each earning a perfect outcome score. This indicates that LLMs can generate financially effective decisions. However, their reasoning quality scores show significant variance, with notable weaknesses in event grounding. The results confirm that achieving profitability does not guarantee logically coherent or well-explained investment theses, highlighting a key area for future improvement.

\begin{table*}
\centering
\caption{Comprehensive Evaluation of LLMs on \textsc{InvestLogicBench2026}}
\label{tab:comprehensive_results}
\begin{tabular}{@{}lcccccc@{}}
\toprule
\multirow{2}{*}{\textbf{Model}} & \multicolumn{2}{c}{\textbf{Reasoning Quality}} & \multicolumn{2}{c}{\textbf{Decision Outcome}} & \multirow{2}{*}{\textbf{Overall Score /10}} \\
\cmidrule(lr){2-3} \cmidrule(lr){4-5}
 & \textbf{Event Coherence (/5)} & \textbf{Logical Plausibility (/5)} & \textbf{Win Rate} & \textbf{Total Profit} & \\
\midrule
DeepSeek-V3 & 2.8 & 4.1 & 72\% & +16.7\% & 7.8 \\
Qwen3-Max & 0.8 & 4.2 & 51\% & +7.7\% & 5.7 \\
Claude-4.5 & 2.2 & 3.9 & 46\% & +5.4\% & 5.3 \\
GPT-5 & 1.8 & 4.1 & 40\% & +3.8\% & 3.1 \\
\bottomrule
\end{tabular}
\vspace{4pt}
\footnotesize \textit{Note: Logical Plausibility evaluates the coherence of the E→R→D chain. Total Profit is the average return of each outcome.}
\end{table*}

\subsection{Case Study: Event-Driven Reasoning Failure}
We present a simulated case where an agent correctly predicted a rally in semiconductor stocks. The agent's logic wove together nuanced observations from earnings call transcripts, industry capacity reports, and geopolitical news. When leading LLMs were provided the same raw event data, they either fixated on the most headline-grabbing news piece, produced a generic "strong demand" summary, or generated internally contradictory statements. This case exemplifies the \textbf{event-noise discrimination} and \textbf{integrative reasoning} deficits of current models.
\begin{figure}[t]
    \centering
    \includegraphics[width=0.5\textwidth]{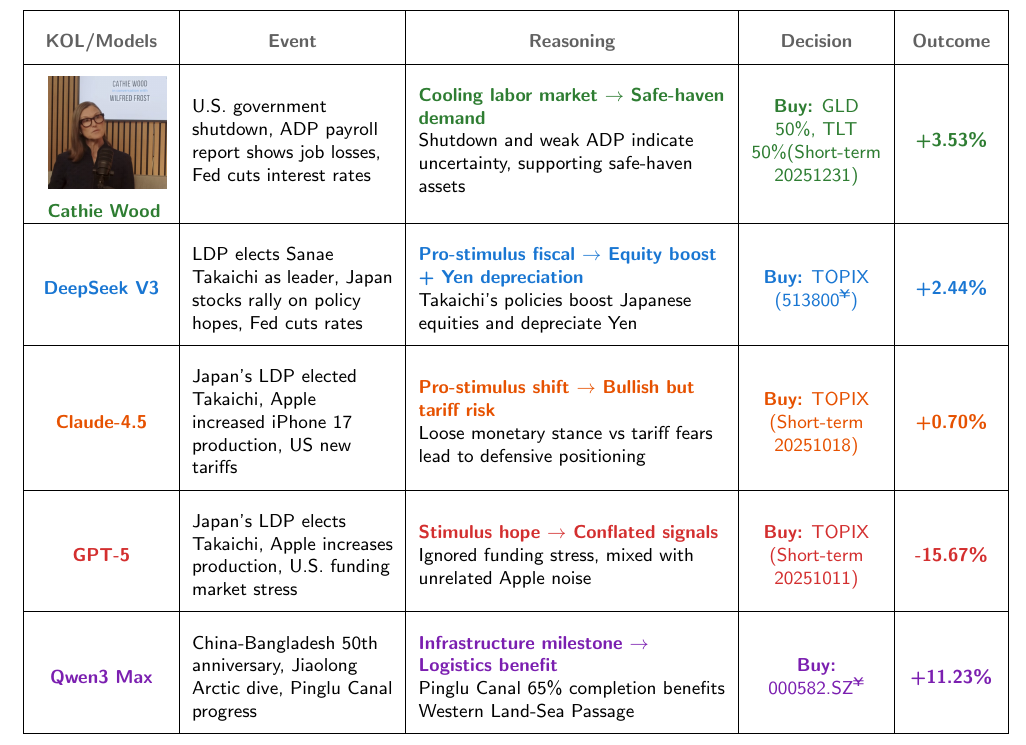}
    \vspace{-0.7cm}
\caption{Decomposing the \textbf{Investment Logic Gap}: A comparative case study of investment reasoning across the \textbf{P$\rightarrow$E$\rightarrow$R$\rightarrow$D$\rightarrow$O chain} between a human expert (Cathie Wood) and LLMs instantiated with her agent profile. The analysis focuses on the market event cluster of October 4, 2025 with empty starting position, and the performance divergence is substantial: the expert returned +3.53\%, the best LLM (Qwen) +11.23\%, and the worst LLM (GPT) -15.67\%, resulting in a 26.9\% LLM performance gap. According to the table, LLMs exhibit significantly divergent investment logic and outcome. We use \textsuperscript{\textyen} to present security code from Chinese financial market. Notably, the distinct geographical bias in asset recommendations across different LLMs reflects inherent market preferences shaped by their training data.}
\label{fig:case_study}
\end{figure}

Figure~\ref{fig:case_study} illustrates a simulated event-driven investment scenario where Cathie Wood correctly predicted a rally in semiconductor stocks by integrating nuanced observations from multiple data sources. In contrast, LLMs exhibit systematic reasoning failures when provided the same raw event data.

\textbf{Key Observations:}
\begin{enumerate}[label=(\arabic*), itemsep=0pt, topsep=2pt]
    \item Human experts excel at \textbf{integrative reasoning}, weaving together earnings call transcripts, industry capacity reports, and geopolitical news into a coherent investment thesis.
    \item LLMs display poor \textbf{event-noise discrimination}, either fixating on headline-grabbing news (e.g., government shutdown) or producing generic summaries lacking specificity.
    \item The quantitative evaluation reveals a 25.7\% gap in overall logic quality and a 2.2-point deficit in noise discrimination between human experts and the best-performing LLM (Claude-4.5-Sonnet).
\end{enumerate}

\textbf{Implications:} This case exemplifies the core limitations identified by \textsc{InvestLogicBench2026}: LLMs struggle to distinguish signal from noise and fail to maintain logical consistency across multiple data sources, explaining their underwhelming performance in live trading environments despite strong performance on static financial benchmarks.

\section{Related Work}
\label{sec:related_work}

LLM evaluation in finance has followed two main tracks: \textit{static benchmarks} testing analytical knowledge, and \textit{dynamic agent frameworks} measuring trading performance. However, a critical gap remains in evaluating the \textit{investment reasoning process} itself---the logical synthesis of unstructured events into actionable theses. Recent studies expose temporal awareness failures and other reasoning limitations \citep{sehgal2026real}, underscoring the need for process-centric evaluation. \textsc{InvestLogicBench2026} addresses this gap by introducing the first benchmark grounded in explicit \textbf{P$\rightarrow$E$\rightarrow$R$\rightarrow$D$\rightarrow$O} decision chains, enabling \textbf{process-diagnostic} assessment of investment logic.

\subsection{Static Benchmarks: Knowledge and Analysis}
\label{subsec:static_benchmarks}

Early work evaluates LLMs as analysts through static QA and numerical reasoning. FinQA \citep{chen2022finqa} and TAT-QA \citep{zhu2021tatqa} require multi-step reasoning over financial reports and hybrid tabular-textual content. Subsequent benchmarks expand scope: FinanceBench \citep{islam2023financebench} tests open-domain QA on SEC filings, while BizBench \citep{koncelkedziorski2024bizbench} reframes business problems as quantitative tasks. Comprehensive suites like FinBen \citep{xie2024finben} aggregate these tasks.

The key limitation is \textit{retrospective analysis} of given information. As noted by \citet{ding2024survey}, strong performance reflects calculation proficiency but not \textit{prospective decision-making} under uncertainty---the core investor function. This limitation persists even in specialized models like BloombergGPT \citep{wu2023bloomberggpt} and FinGPT \citep{yang2023fingpt}, which remain predominantly evaluated on static benchmarks.

\subsection{Dynamic Agents: Real-World Performance}
\label{subsec:dynamic_agents}

Recent work shifts from evaluating models as tools to assessing them as autonomous \textit{agents} in simulated or live markets. This paradigm prioritizes outcome-based metrics (P\&L). Frameworks like FinDABench \citep{liu2024findabench} and XFinBench \citep{zhang2025xfinbench} place LLM agents in complex financial problem-solving environments. Fully autonomous arenas such as AI-Trader \citep{fan2025aitrader} and DeepFund \citep{li2025deepfund} use real-time data with actual capital, providing high ecological validity. Architecturally, this space features multi-agent systems \citep{yu2024fincon} and diverse agent-based modeling approaches \citep{gao2023largelanguagemodels}.

While crucial for end-to-end stress-testing, these approaches are fundamentally \textbf{outcome-opaque}. They indicate \textit{whether} an agent profits but not \textit{why}. Profitable trades may stem from sound logic or noise; losses may reflect poor strategy or black-swan events. This black-box nature hinders targeted improvement and obscures the root cause of the ``Capabilities-Performance Paradox'' \citep{ding2024survey}.

\subsection{Bridging the Gap: Process-Diagnostic Evaluation}
\label{subsec:process_diagnostic}

To address these limitations, recent work diagnoses specific reasoning failures. Look-ahead bias remains a critical challenge, causing performance decay under point-in-time constraints. Complementary studies examine temporal reasoning under pressure \citep{sehgal2026real} and advanced market analysis using LLM agents \citep{fatouros2025marketsense}.

Most pertinent is the \textbf{Chain-of-Decision} framework by \citet{chen2024distilling}, which argues that modeling professional decisions requires distilling explicit intermediate reasoning steps between information intake and actions. This directly informs our \textbf{P$\rightarrow$E$\rightarrow$R$\rightarrow$D$\rightarrow$O} schema. However, their work focuses on report generation rather than the full investment cycle encompassing decision, action, and market-validated outcome.

Unlike prior work measuring input knowledge or output performance, we evaluate the \textit{reasoning chain} connecting them by (1) \textbf{grounding evaluation in real expert decision trails}, (2) \textbf{assessing logical coherence independent of P\&L}, and (3) \textbf{enabling style-conditioned diagnosis}. This provides the substrate to quantify the investment logic gap and align models toward robust, interpretable, expert-aligned financial reasoning.

\section{Conclusion}
This paper identifies the absence of benchmarks for \textit{practical investment logic}—the reasoning chain—as a fundamental obstacle to the development of investment-oriented agents. We introduce \textsc{InvestLogicBench2026}, the first benchmark constructed from the event-driven reasoning traces of real-world investors. It provides structured, validated data to evaluate an LLM's ability to formulate and maintain coherent investment theses. Our analysis confirms a significant \textit{investment logic gap}: state-of-the-art LLMs fail to match the nuanced, contextual reasoning of human experts, explaining their poor performance in live trading. This reproducible benchmark shifts the evaluation paradigm from passive knowledge recall to active reasoning. It lays the essential foundation for future work in training logic-aligned models and architecting personalized, market-aligned investment agents.

\section{Impact Statement}
\label{sec:impact}

This work introduces \textsc{InvestLogicBench2026}, a benchmark designed to advance machine learning by evaluating the core capability of dynamic, sequential reasoning under uncertainty—essential for developing reliable autonomous agents in finance.

By shifting evaluation from opaque performance metrics to the diagnosis of reasoning quality, the benchmark facilitates the development of trustworthy, interpretable, and logically sound AI systems. It serves as a foundational step toward transparent and accountable decision-making AI, with broader implications for improving financial decision-support tools.

\bibliographystyle{plainnat}
\bibliography{references}

@article{ding2024survey,
  title={Large Language Model Agent in Financial Trading: A Survey},
  author={Ding, Han and Li, Yinheng and Wang, Junhao and Chen, Hang},
  journal={arXiv preprint arXiv:2408.06361},
  year={2024},
  url={https://arxiv.org/abs/2408.06361}
}

@article{chen2022finqa,
  title={Fin{QA}: A Dataset of Numerical Reasoning over Financial Data},
  author={Chen, Zhiyu and Chen, Wenhu and Smiley, Charese and Shah, Sameena and Borova, Iana and Langdon, Dylan and Moussa, Reema and Beane, Matt and Huang, Ting-Hao and Routledge, Bryan and Wang, William Yang},
  journal={arXiv preprint arXiv:2109.00122},
  year={2022},
  url={https://arxiv.org/abs/2109.00122}
}

@article{zhu2021tatqa,
  title={{TAT-QA}: A Question Answering Benchmark on a Hybrid of Tabular and Textual Content in Finance},
  author={Zhu, Fengbin and Lei, Wenqiang and Huang, Youcheng and Wang, Chao and Zhang, Shuo and Lv, Jiancheng and Feng, Fuli and Chua, Tat-Seng},
  journal={arXiv preprint arXiv:2105.07624},
  year={2021},
  url={https://arxiv.org/abs/2105.07624}
}

@article{islam2023financebench,
  title={Finance{B}ench: A New Benchmark for Financial Question Answering},
  author={Islam, Pranab and Kannappan, Anand and Kiela, Douwe and Qian, Rebecca and Scherrer, Nino and Vidgen, Bertie},
  journal={arXiv preprint arXiv:2311.11944},
  year={2023},
  url={https://arxiv.org/abs/2311.11944}
}

@article{koncelkedziorski2024bizbench,
  title={{B}iz{B}ench: A Quantitative Reasoning Benchmark for Business and Finance},
  author={Koncel-Kedziorski, Rik and Krumdick, Michael and Lai, Viet and Reddy, Varshini and Lovering, Charles and Tanner, Chris},
  journal={arXiv preprint arXiv:2311.06602},
  year={2024},
  url={https://arxiv.org/abs/2311.06602}
}

@inproceedings{xie2024finben,
  title={FinBen: A Holistic Financial Benchmark for Large Language Models},
  author={Xie, Qianqian and Wang, Weiran and Chen, Zhenyu and Xiang, Ruoyu and Zhang, Xiao and Deng, Yansong and others},
  booktitle={Advances in Neural Information Processing Systems 38 (NeurIPS 2024)},
  year={2024},
  url={https://proceedings.neurips.cc/paper_files/paper/2024/hash/adb1d9fa8be4576d28703b396b82ba1b-Abstract-Datasets_and_Benchmarks_Track.html},
  doi={10.52202/079017-3033}
}

@inproceedings{liu2024findabench,
  title={Fin{DAB}ench: Benchmarking Financial Data Analysis Ability of Large Language Models},
  author={Liu, Shu and Zhao, Shangqing and Jia, Chenghao and Zhuang, Xinlin and Long, Zhaoguang and Zhou, Jie and Zhou, Aimin and Lan, Man and Chong, Yang},
  booktitle={Proceedings of the International Conference on Computational Linguistics},
  year={2024},
  url={https://api.semanticscholar.org/CorpusID:266844370}
}

@article{zhang2025xfinbench,
  title={{XFinBench}: Benchmarking {LLM}s in Complex Financial Problem Solving and Reasoning},
  author={Zhang, Zhihan and Cao, Yixin and Liao, Lizi},
  journal={arXiv preprint arXiv:2508.15861},
  year={2025},
  url={https://arxiv.org/abs/2508.15861}
}

@article{wu2023bloomberggpt,
  title={{B}loomberg{GPT}: A Large Language Model for Finance},
  author={Wu, Shijie and Irsoy, Ozan and Lu, Steven and Dabravolski, Vadim and Dredze, Mark and Gehrmann, Sebastian and Kambadur, Prabhanjan and Rosenberg, David and Mann, Gideon},
  journal={arXiv preprint arXiv:2303.17564},
  year={2023},
  url={https://arxiv.org/abs/2303.17564}
}

@article{yang2023fingpt,
  title={Fin{GPT}: Open-Source Financial Large Language Models},
  author={Yang, Hongyang and Liu, Xiao-Yang and Wang, Christina Dan},
  journal={arXiv preprint arXiv:2306.06031},
  year={2023},
  url={https://arxiv.org/abs/2306.06031}
}

@article{fan2025aitrader,
  title={{AI}-Trader: Benchmarking Autonomous Agents in Real-Time Financial Markets},
  author={Fan, Tianyu and Yang, Yuhao and Jiang, Yangqin and Zhang, Yifei and Chen, Yuxuan and Huang, Chao},
  journal={arXiv preprint arXiv:2512.10971},
  year={2025},
  url={https://arxiv.org/abs/2512.10971}
}

@article{li2025deepfund,
  title={Will {LLM}s be Professional at Fund Investment? {D}eep{F}und: A Live Arena Perspective},
  author={Li, Changlun and Shi, Yao and Luo, Yuyu and Tang, Nan},
  journal={arXiv preprint arXiv:2503.18313},
  year={2025},
  url={https://arxiv.org/abs/2503.18313}
}

@article{chen2024distilling,
  title={Distilling Analysis from Generative Models for Investment Decisions},
  author={Chen, Chung-Chi and Takamura, Hiroya and Kobayashi, Ichiro and Miyao, Yusuke},
  journal={arXiv preprint arXiv:2410.07225},
  year={2024},
  url={https://arxiv.org/abs/2410.07225}
}

@article{sehgal2026real,
  title={Real-Time Deadlines Reveal Temporal Awareness Failures in {LLM} Strategic Dialogues},
  author={Sehgal, Neil K. R. and Guntuku, Sharath Chandra and Ungar, Lyle},
  journal={arXiv preprint arXiv:2601.13206},
  year={2026},
  url={https://arxiv.org/abs/2601.13206}
}

@inproceedings{yu2024fincon,
  title={Fin{C}on: A Synthesized {LLM} Multi-Agent System with Conceptual Verbal Reinforcement for Enhanced Financial Decision Making},
  author={Yu, Yangyang and Yao, Zhiyuan and Li, Haohang and Deng, Zhiyang and Jiang, Yuechen and Cao, Yupeng and Chen, Zhi and Suchow, Jordan W. and Cui, Zhenyu and Liu, Rong and Xu, Zhaozhuo and Subbalakshmi, Koduvayur and Xiong, Guojun and He, Yueru and Huang, Jimin and Li, Dong and Xie, Qianqian},
  booktitle={Advances in Neural Information Processing Systems},
  volume={38},
  year={2024}
}

@article{fatouros2025marketsense,
  title={{M}arket{S}ense{AI} 2.0: Enhancing Stock Analysis through {LLM} Agents},
  author={Fatouros, George and Metaxas, Kostas and Soldatos, John and Karathanassis, Manos},
  journal={arXiv preprint arXiv:2502.00415},
  year={2025},
  url={https://arxiv.org/abs/2502.00415}
}

@article{gao2023largelanguagemodels,
  title={Large Language Models Empowered Agent-based Modeling and Simulation: A Survey and Perspectives},
  author={Gao, Chen and Lan, Xiaochong and Li, Nian and Yuan, Yuan and Ding, Jingtao and Zhou, Zhilun and Xu, Fengli and Li, Yong},
  journal={arXiv preprint arXiv:2312.11970},
  year={2023},
  url={https://arxiv.org/abs/2312.11970}
}

\clearpage
\onecolumn

\appendix
\section{Appendix}

\subsection{KOL Sample Construction and Measurement Notes}
\label{app:sample}
\noindent
This appendix documents a curated sample of YouTube-based key opinion leaders (KOLs) for descriptive comparison of \emph{investment-logic-oriented} content.
Channels are selected based on whether their content primarily articulates \textbf{generalizable decision frameworks} (e.g., macro--industry--firm reasoning pipelines; valuation, risk management, and portfolio construction principles; cycle and sentiment indicators; and rule-based trading systems), as opposed to predominantly event-driven or news-reactive commentary.
The resulting sample covers multiple market segments (equities/funds/macro, options and trading education, and crypto/Web3), enabling cross-domain characterization of audience scale and content supply.

\par\medskip

\noindent
As shown in Table~\ref{tab:kol_stats}, \textbf{Subscribers} is used as a coarse proxy for audience size, and \textbf{Videos} reflects cumulative content output and, to a limited extent, channel maturity.
Both measures are time-varying; thus, the table should be interpreted as a descriptive snapshot rather than a fixed ground-truth record.
For subsequent quantitative analyses, we recommend re-scraping and freezing all statistics within a single, pre-specified time window to ensure comparability.

\par\bigskip

\renewcommand{\arraystretch}{1.5}
\setlength{\tabcolsep}{4pt}
\footnotesize

\begin{longtable}{@{} p{0.25\textwidth} >{\raggedright\arraybackslash}p{0.53\textwidth} p{0.10\textwidth} p{0.07\textwidth} @{}}
\caption{\textbf{Statistics of Selected KOL Creators}}
\label{tab:kol_stats} \\
\toprule
\textbf{Name} & \textbf{Description} & \textbf{Subscribers} & \textbf{Videos} \\
\midrule
\endfirsthead

\toprule
\textbf{Name} & \textbf{Description} & \textbf{Subscribers} & \textbf{Videos} \\
\midrule
\endhead

Cathie Wood & Investment strategist, founder of ARK Invest & 619K & 743 \\
Zip Trader (Charlie) & Day trading and swing trading education & 857 & 819 \\
Stock Moe & Stock and crypto trading strategies & 714K & 897 \\
Data Dispatch & Market data and analytics insights & 28.4K & 611 \\
Paul Thomas Investing & Penny stocks and growth investment analysis & 57.4K & 478 \\
Stas Talks Stocks & Stock market discussion and analysis & 61.6K & 4283 \\
Lark Davis & Crypto investment strategies & 638K & 4936 \\
Miles Deutscher & Cryptocurrency analysis and portfolio strategies & 224K & 2284 \\
Financial Education & Investing education & 911K & 2800 \\
Everything Money & Investing & 376K & 3019 \\
Joseph Carlson Show & Investing & 499K & 449 \\
Learn to Invest & Investing basics & 289K & 597 \\
Sven Carlin & Value investing & 263K & 2176 \\
New Money & Finance commentary & 1090K & 415 \\
Tom Nash & Market analysis & 572K & 880 \\
Chicken Genius Singapore & Singapore investing & 249K & 249 \\
Ticker Symbol YOU & Investing & 572K & 382 \\
Jimmy Dividend Investing & Dividend investing & 289K & 597 \\
TraderTV Live & Professional live market trading sessions & 550K & 6315 \\
The Plain Bagel & Animated financial education & 1140K & 263 \\
The Compound & Casual market commentary & 223K & 1885 \\
Seeking Alpha & Fundamental stock analysis & 32.7K & 947 \\
Patrick Boyle & Academic finance perspective & 1100K & 460 \\
Option Alpha & Systematic options trading education & 296K & 862 \\
NaNa Talks US Stocks & Chinese-language US stock analysis & 301K & 1573 \\
Morningstar & Fund analysis and long-term strategies & 129K & 2956 \\
Money or Life & Financial freedom philosophy & 71.8K & 217 \\
Investopedia & Financial dictionary and concepts & 298K & 545 \\
Investor's Business Daily & CAN SLIM investing system & 169K & 4961 \\
Hedgeye & Macroeconomic analysis & 78.3K & 4131 \\
Coin Bureau & Comprehensive cryptocurrency education & 2730K & 1773 \\
The Defiant & DeFi and Web3 news and analysis & 131K & 1346 \\
Unchained & Blockchain and crypto industry insights & 6770K & 1649 \\
Crypto Casey & Beginner-friendly cryptocurrency education & 678K & 875 \\
The Maverick of Wall Street & Alternative investment strategies & 146K & 1686 \\
Larry Jones & Stocks and crypto analysis for entrepreneurs & 691K & 2154 \\
Meet Kevin & Real estate and financial independence education & 2050K & 6095 \\
Graham Stephan & Personal finance and real estate investing & 5140K & 1402 \\
Mark Moss & Economic trends and investment strategies & 803K & 1365 \\
Real Vision & In-depth financial markets analysis & 89.7K & 1210 \\
Garys Economics & Economic analysis and market commentary & 1520K & 369 \\
Let's Talk Money! & Personal finance education and tips & 738K & 1199 \\
ClearValue Tax & Tax strategies and financial planning & 2820K & 151 \\
Kitco NEWS & Precious metals market news and analysis & 750K & 4818 \\
\bottomrule
\end{longtable}
\clearpage
\onecolumn

\section{Appendix: KOL Resources}
\label{app:kol_urls}

\subsection{KOL Channel URLs and Reproducibility}
\noindent
To support reproducibility and facilitate follow-up analyses, Table~\ref{tab:kol_urls} provides the canonical channel URLs for the curated KOL sample reported in Table~\ref{tab:kol_stats}.
These links enable consistent access to each source and allow downstream data collection (e.g., channel metadata, upload cadence, topical coverage, and representative playlists) under a fixed protocol.
In empirical settings where platform statistics and content inventories evolve over time, anchoring the sample with persistent URLs is essential for re-scraping within a pre-specified time window and ensuring that subsequent measurements remain comparable across runs.

\par\bigskip

\begin{longtable}{@{} p{0.35\textwidth} p{0.6\textwidth} @{}}
\caption{\textbf{List of KOL Channels and URLs}}
\label{tab:kol_urls} \\
\toprule
\textbf{KOL Name} & \textbf{Channel URL} \\
\midrule
\endfirsthead

\toprule
\textbf{KOL Name} & \textbf{Channel URL} \\
\midrule
\endhead

Cathie Wood & \url{https://www.youtube.com/@ARKInvest2015} \\
Financial Education & \url{https://www.youtube.com/@FinancialEducation} \\
Everything Money & \url{https://www.youtube.com/@EverythingMoney} \\
Joseph Carlson Show & \url{https://www.youtube.com/@JosephCarlsonShow} \\
Learn to Invest & \url{https://www.youtube.com/@LearnToInvest} \\
Sven Carlin & \url{https://www.youtube.com/c/InvestwithSvenCarlinPhD} \\
New Money & \url{https://www.youtube.com/@NewMoneyYouTube} \\
Tom Nash & \url{https://www.youtube.com/@TomNashTV} \\
Chicken Genius Singapore & \url{https://www.youtube.com/@ChickenGeniusSingapore} \\
Ticker Symbol YOU & \url{https://www.youtube.com/@TickerSymbolYOU} \\
Jimmy Dividend Investing & \url{https://www.youtube.com/@JimmyDividend} \\
TraderTV Live & \url{https://www.youtube.com/c/TraderTVLive} \\
The Plain Bagel & \url{https://www.youtube.com/c/ThePlainBagel} \\
The Compound & \url{https://www.youtube.com/@TheCompoundNews} \\
Seeking Alpha & \url{https://www.youtube.com/c/SeekingAlpha} \\
Patrick Boyle & \url{https://www.youtube.com/c/PatrickBoyleOnFinance} \\
Option Alpha & \url{https://www.youtube.com/c/OptionAlpha} \\
NaNa Talks US Stocks & \url{https://www.youtube.com/@NaNaShuoMeiGu} \\
Morningstar & \url{https://www.youtube.com/@morningstar} \\
Money or Life & \url{https://www.youtube.com/@Money_or_Life} \\
Investopedia & \url{https://www.youtube.com/c/Investopedia} \\
Investor's Business Daily & \url{https://www.youtube.com/c/InvestorsBusinessDaily} \\
Hedgeye & \url{https://www.youtube.com/user/Hedgeye} \\
Zip Trader (Charlie) & \url{https://www.youtube.com/c/ziptrader} \\
Stock Moe & \url{https://www.youtube.com/channel/UCoMzWLaPjDJBbipihD694pQ} \\
Data Dispatch & \url{https://www.youtube.com/@DataDispatch} \\
Paul Thomas Investing & \url{https://www.youtube.com/c/PaulThomasInvesting} \\
Stas Talks Stocks & \url{https://www.youtube.com/@StasTalksStocks} \\
Lark Davis & \url{https://www.youtube.com/@TheCryptoLark} \\
Miles Deutscher & \url{https://www.youtube.com/@MilesDeutscher} \\
Coin Bureau & \url{https://www.youtube.com/@CoinBureau} \\
The Defiant & \url{https://www.youtube.com/@TheDefiant} \\
Unchained & \url{https://www.youtube.com/@UnchainedOff} \\
Crypto Casey & \url{https://www.youtube.com/@CryptoCasey} \\
The Maverick of Wall Street & \url{https://www.youtube.com/@TheMaverickofWallStreet} \\
Larry Jones & \url{https://www.youtube.com/c/STOCKUPwithLarryJones} \\
Meet Kevin & \url{https://www.youtube.com/@MeetKevin} \\
Graham Stephan & \url{https://www.youtube.com/@GrahamStephan} \\
Mark Moss & \url{https://www.youtube.com/@1MarkMoss} \\
Real Vision & \url{https://www.youtube.com/@RealVisionFinance} \\
Garys Economics & \url{https://www.youtube.com/@garyseconomics} \\
Let's Talk Money! & \url{https://www.youtube.com/@josephhogue} \\
ClearValue Tax & \url{https://www.youtube.com/@clearvaluetax9382} \\
Kitco NEWS & \url{https://www.youtube.com/@kitco} \\
\bottomrule
\end{longtable}

\twocolumn


\clearpage
\onecolumn

\section{Appendix:KOL Fund Reference Tabless}
\label{app:kol_resource}

\subsection{Institutions, Fund Managers, and Morningstar Ratings}
\label{app:kol_ms_ratings}

\noindent
Table~\ref{tab:firm_manager_fund_ms} reports a compact snapshot of the fund universe by listing each record’s institution (fund firm), fund manager(s), fund name, and the Morningstar \emph{overall} star rating.
We include the star rating as a standardized, coarse descriptor of historical risk-adjusted performance within a fund’s peer group, and use it strictly for descriptive comparison rather than causal interpretation.
Because manager assignments and ratings may change over time, the table should be interpreted as a snapshot tied to the data extraction window.

\par\bigskip

\renewcommand{\arraystretch}{1.2}
\setlength{\tabcolsep}{4pt}
\footnotesize

\begin{longtable}{@{}%
>{\raggedright\arraybackslash}p{0.18\textwidth}
>{\raggedright\arraybackslash}p{0.28\textwidth}
>{\raggedright\arraybackslash}p{0.38\textwidth}
c
@{}}
\caption{\textbf{Institutions, Fund Managers, Fund Names, and Morningstar Overall Ratings}}
\label{tab:firm_manager_fund_ms} \\
\toprule
\textbf{Institution (Firm)} & \textbf{Fund Manager(s)} & \textbf{Fund Name} & \textbf{MS Rating} \\
\midrule
\endfirsthead

\toprule
\textbf{Institution (Firm)} & \textbf{Fund Manager(s)} & \textbf{Fund Name} & \textbf{MS Rating} \\
\midrule
\endhead

Aegis & Scott L. Barbee & Aegis Value I & 5 \\
Alger & Patrick Kelly, Ankur Crawford & Alger Focus Equity Y & 5 \\
Alger & Gregory S. Adams & Alger Growth \& Income Z & 4 \\
Allspring Global Investments & Justin P. Carr, Robert M. Wicentowski & Allspring Disciplined US Core R6 & 5 \\
Allspring Global Investments & John R. Campbell, Vincent Fioramonti & Allspring Large Cap Core R6 & 5 \\
Gotham & Joel Greenblatt, Robert Goldstein & Gotham Index Plus Institutional & 5 \\
American Century Investments & Keith Lee, Jeffrey R. Bourke, Tong Li & American Century U.S. Equity Focus I & 3 \\
BNY Mellon & Peter D. Goslin & BNY Mellon Appreciation Instl & 5 \\
Ave Maria Mutual Funds & George P. Schwartz, Timothy S. Schwartz & Ave Maria Rising Dividend I & 5 \\
VALIC & Jeffrey D. Parker, Brian Demain, Philip Cody White & VALIC Company I International Growth & 5 \\
WesMark & Douglas C. Robertson & WesMark Small Cap I & 4 \\
Wasatch & Garrett S. Lamb, Sean A. Schaub & Wasatch Small Cap Value Inst & 4 \\
Hood River Capital Management & Michael J. Delf, John H. Beck & Hood River Small-Cap Growth I & 5 \\
GMO & Ben Inker, Jeremy Grantham & GMO Quality Fund III & 5 \\
Leavell & Thomas A. Rowe, Ryan T. Leavell & Leavell Investment Counsel Mid Cap Growth & 5 \\
SEI & SEI Investments Management Corporation & SEI Large Cap Growth Instl (SIMT) & 4 \\
T. Rowe Price & Larry J. Puglia & T. Rowe Price Blue Chip Growth I & 4 \\
T. Rowe Price & Justin P. White & T. Rowe Price Growth Stock I & 4 \\
T. Rowe Price & Brian W. Berghuis & T. Rowe Price Mid-Cap Growth I & 4 \\
T. Rowe Price & Brian C. Rogers & T. Rowe Price Equity Income I & 4 \\
BlackRock & Robert R. Shillito, Michael C. Clemente & BlackRock Advantage Small Cap Core Instl & 4 \\
BlackRock & Tony DeSpirito, John M. H. Lawler, Kevin B. Winkler & BlackRock Global Equity Instl & 4 \\
BlackRock & Tony DeSpirito, John M. H. Lawler, Kevin B. Winkler & BlackRock Mid-Cap Growth Equity Instl & 4 \\
BlackRock & Tony DeSpirito, John M. H. Lawler, Kevin B. Winkler & BlackRock Small Cap Growth Equity Instl & 4 \\
Franklin Templeton Investments & David M. Winters, Todd Brighton & Franklin Small Cap Growth Adv & 4 \\
Copley & Jeffrey M. Fischer, Benjamin W. Hall & Copley Focus Equity Institutional & 4 \\
Hotchkis \& Wiley & Daniel G. Hotchkis, Shannon P. Kress & Hotchkis \& Wiley Value Opportunities I & 4 \\
Independent Franchise Partners & Andrew L. Moffett, David W. Potter & Independent Franchise Partners US Equity I & 4 \\
Fidelity Investments & Will Danoff & Fidelity Contrafund K6 & 5 \\
Fidelity Investments & Joel P. Tillinghast & Fidelity Low-Priced Stock K6 & 4 \\
Fidelity Investments & Geode Capital Management & Fidelity Total Market Index & 4 \\
Dodge \& Cox & Dodge \& Cox & Dodge \& Cox Income & 4 \\
Dodge \& Cox & Dodge \& Cox & Dodge \& Cox Stock & 4 \\
ClearBridge Investments & David C. Kupfer, Peter W. Fondren & ClearBridge Large Cap Growth IS & 4 \\
ClearBridge Investments & Robert E. Taylor, John D. Freeman & ClearBridge Large Cap Value IS & 4 \\
Cohen \& Steers & Joseph M. Harvey, Jason T. Yablon & Cohen \& Steers Realty Shares Instl & 3 \\
Columbia Threadneedle Investments & James P. Tierney & Columbia Seligman Global Technology Instl 2 & 5 \\
Dimensional Fund Advisors & Dimensional Fund Advisors & DFA Emerging Markets Core Equity I & 4 \\
Dimensional Fund Advisors & Dimensional Fund Advisors & DFA International Core Equity I & 4 \\
Dimensional Fund Advisors & Dimensional Fund Advisors & DFA US Core Equity 1 I & 4 \\
Invesco & Amit D. Kumar & Invesco Developing Markets Instl & 3 \\
J.P. Morgan & Clare H. Hart, Michael D. Marini & JPMorgan Equity Income R6 & 4 \\
J.P. Morgan & Jonathan D. Simon, Jennifer L. Williams & JPMorgan Large Cap Growth R6 & 4 \\
MFS Investment Management & Kevin D. Clark, Joseph Flaherty & MFS Growth R6 & 4 \\
MFS Investment Management & Robert M. Zito, James H. LeBoeuf & MFS Value R6 & 4 \\
PIMCO & Scott A. Mather, Esteban J. Burbano & PIMCO Total Return Instl & 3 \\
Principal & William R. Bowen, Allen E. Jackson & Principal MidCap R6 & 4 \\
Charles Schwab & Schwab Equity Ratings & Schwab S\&P 500 Index & 4 \\
Artisan Partners & David Samra, Kiki P. Pillai & Artisan Developing World Instl & 4 \\
Artisan Partners & Benjamin Herrick, David Samra, Kiki P. Pillai & Artisan International Value Instl & 4 \\
Baron Capital & Ronald Baron, Michael J. Lippert & Baron Partners Instl & 4 \\
Baron Capital & Ronald Baron, Michael T. Sheehy, Neal E. Clark & Baron Small Cap Instl & 4 \\
Vanguard & Vanguard Equity Index Group & Vanguard 500 Index Admiral & 4 \\
Vanguard & Donald M. Kilbride, Michael D. Reckmeyer & Vanguard Dividend Growth Investor & 4 \\
Vanguard & Vanguard Equity Index Group & Vanguard Extended Market Index Admiral & 4 \\
Vanguard & Vanguard Equity Index Group & Vanguard FTSE All-World ex-US Index Admiral & 4 \\
Vanguard & Vanguard Equity Index Group & Vanguard Growth Index Admiral & 4 \\
Vanguard & Vanguard Equity Index Group & Vanguard Small-Cap Index Admiral & 4 \\
Vanguard & Vanguard Equity Index Group & Vanguard Total Stock Market Index Admiral & 4 \\
Vanguard & Vanguard Equity Index Group & Vanguard Value Index Admiral & 4 \\
Vanguard & John T. Keogh, Thomas J. Martin & Vanguard Wellington Investor & 4 \\
Vanguard & John T. Keogh, Thomas J. Martin & Vanguard Windsor II Investor & 4 \\
Vanguard & John T. Keogh, Thomas J. Martin & Vanguard Windsor Investor & 4 \\
Vanguard & John T. Keogh, Thomas J. Martin & Vanguard Wellesley Income Investor & 4 \\
Vanguard & Vanguard Equity Index Group & Vanguard World Stock Index Investor & 4 \\

\bottomrule
\end{longtable}
\twocolumn

\clearpage
\onecolumn

\subsection{Fund Managers and Trailing Return Metrics}
\label{app:kol_trailing_returns}

\noindent
Table~\ref{tab:fund_mgr_returns} reports a compact summary of each fund’s manager(s) and trailing performance metrics.
We use \textbf{TR 1Y} to denote the trailing one-year total return, and \textbf{TR 3Y} / \textbf{TR 5Y} to denote the annualized total returns over the trailing three- and five-year horizons, respectively.
These values are time-varying and should be interpreted as a descriptive snapshot tied to the data extraction window.

\par\bigskip

\sisetup{
  detect-all,
  table-number-alignment = center,
  round-mode = places,
  round-precision = 2
}

\renewcommand{\arraystretch}{1.15}
\setlength{\tabcolsep}{3pt}
\footnotesize
\setlength{\LTleft}{0pt}
\setlength{\LTright}{0pt}


\setlength{\tabcolsep}{8pt}
\setlength{\LTleft}{0pt}
\setlength{\LTright}{0pt}

\noindent\hfill
\begin{longtable}{@{}%
  >{\raggedright\arraybackslash}p{0.4\linewidth}
  >{\raggedright\arraybackslash}p{0.26\linewidth}
  @{\hspace{2em}}
  S[table-format=2.2, table-align-text-post=false]
  @{\hspace{2em}}
  S[table-format=2.2, table-align-text-post=false]
  @{\hspace{2em}}
  S[table-format=2.2, table-align-text-post=false]
@{}}
\caption{\textbf{Fund Managers and Trailing Returns}}
\label{tab:fund_mgr_returns} \\
\toprule
\textbf{Fund Name} & \textbf{Manager(s)} &
\multicolumn{1}{c}{\textbf{1Y}} &
\multicolumn{1}{c}{\textbf{3Y}} &
\multicolumn{1}{c}{\textbf{5Y}} \\
\midrule
\endfirsthead

\toprule
\textbf{Fund Name} & \textbf{Manager(s)} &
\multicolumn{1}{c}{\textbf{1Y}} &
\multicolumn{1}{c}{\textbf{3Y}} &
\multicolumn{1}{c}{\textbf{5Y}} \\
\midrule
\endhead

Aegis Value I & S. Barbee & 29.37\% & 16.90\% & 24.01\% \\
Alger Focus Equity Y & P. Kelly, A. Crawford & 51.38\% & 32.67\% & 19.60\% \\
Alger Growth \& Income Z & G. Adams & 14.82\% & 16.25\% & 16.23\% \\
Allspring Disciplined US Core R6 & J. Carr, R. Wicentowski & 18.56\% & 18.61\% & 17.29\% \\
Allspring Large Cap Core R6 & J. Campbell, V. Fioramonti & 19.50\% & 19.34\% & 17.89\% \\
Allspring Special Small Cap Value R6 & J. Campbell, V. Fioramonti & 16.06\% & 12.17\% & 12.77\% \\
American Funds Washington Mutual R6 & M. Pacheco, K. Martell, B. Reick & 10.84\% & 10.86\% & 10.71\% \\
Artisan Developing World Instl & D. Samra, K. Pillai & 10.61\% & 5.73\% & 7.13\% \\
Artisan International Value Instl & B. Herrick, D. Samra, K. Pillai & 9.68\% & 6.28\% & 10.21\% \\
Baron Partners Instl & R. Baron, M. Lippert & 37.09\% & 17.08\% & 19.25\% \\
Baron Small Cap Instl & R. Baron, M. Sheehy, N. Clark & 17.52\% & 10.71\% & 12.45\% \\
BlackRock Advantage Small Cap Core Instl & R. Shillito, M. Clemente & 17.89\% & 10.62\% & 11.57\% \\
BlackRock Global Equity Instl & T. DeSpirito, J. Lawler, K. Winkler & 24.73\% & 12.81\% & 13.42\% \\
BlackRock Mid-Cap Growth Equity Instl & T. DeSpirito, J. Lawler, K. Winkler & 33.92\% & 15.86\% & 15.05\% \\
BlackRock Small Cap Growth Equity Instl & T. DeSpirito, J. Lawler, K. Winkler & 26.72\% & 10.76\% & 10.50\% \\
ClearBridge Large Cap Growth IS & D. Kupfer, P. Fondren & 35.58\% & 15.01\% & 13.90\% \\
ClearBridge Large Cap Value IS & R. Taylor, J. Freeman & 11.51\% & 10.48\% & 10.64\% \\
Cohen \& Steers Realty Shares Instl & J. Harvey, J. Yablon & 6.58\% & 1.68\% & 4.91\% \\
Columbia Seligman Global Technology Instl 2 & J. Tierney & 46.84\% & 19.32\% & 22.11\% \\
DFA Emerging Markets Core Equity I & Dimensional (DFA) & 8.96\% & 5.62\% & 6.38\% \\
DFA International Core Equity I & Dimensional (DFA) & 13.62\% & 9.40\% & 11.17\% \\
DFA US Core Equity 1 I & Dimensional (DFA) & 18.48\% & 11.49\% & 12.38\% \\
Dodge \& Cox Income & Dodge \& Cox & 1.67\% & 2.57\% & 1.83\% \\
Dodge \& Cox Stock & Dodge \& Cox & 9.53\% & 9.88\% & 10.58\% \\
Fidelity Contrafund K6 & W. Danoff & 27.86\% & 13.63\% & 14.26\% \\
Fidelity Low-Priced Stock K6 & J. Tillinghast & 17.11\% & 9.69\% & 11.32\% \\
Fidelity Total Market Index & Geode & 18.25\% & 11.24\% & 12.28\% \\
Franklin Small Cap Growth Adv & D. Winters, T. Brighton & 28.73\% & 9.78\% & 10.52\% \\
Invesco Developing Markets Instl & A. Kumar & 10.46\% & 4.55\% & 5.59\% \\
JPMorgan Equity Income R6 & C. Hart, M. Marini & 10.24\% & 10.62\% & 11.31\% \\
JPMorgan Large Cap Growth R6 & J. Simon, J. Williams & 36.55\% & 14.63\% & 14.26\% \\
MFS Growth R6 & K. Clark, J. Flaherty & 30.64\% & 12.54\% & 13.26\% \\
MFS Value R6 & R. Zito, J. LeBoeuf & 10.53\% & 9.85\% & 10.79\% \\
PIMCO Total Return Instl & S. Mather, E. Burbano & 1.46\% & 2.38\% & 1.63\% \\
Principal MidCap R6 & W. Bowen, A. Jackson & 25.36\% & 11.83\% & 13.10\% \\
Schwab S\&P 500 Index & Schwab & 18.34\% & 10.99\% & 12.17\% \\
T. Rowe Price Blue Chip Growth I & L. Puglia & 37.78\% & 14.29\% & 14.87\% \\
T. Rowe Price Equity Income I & B. Rogers & 11.05\% & 10.36\% & 10.77\% \\
T. Rowe Price Growth Stock I & J. White & 33.65\% & 12.77\% & 13.26\% \\
T. Rowe Price Mid-Cap Growth I & B. Berghuis & 33.17\% & 12.39\% & 13.07\% \\
Vanguard 500 Index Admiral & Vanguard Index & 18.35\% & 11.00\% & 12.20\% \\
Vanguard Dividend Growth Investor & D. Kilbride, M. Reckmeyer & 8.65\% & 9.91\% & 10.87\% \\
Vanguard Extended Market Index Admiral & Vanguard Index & 19.55\% & 8.72\% & 10.15\% \\
Vanguard FTSE All-World ex-US Index Admiral & Vanguard Index & 13.11\% & 9.43\% & 11.08\% \\
Vanguard Growth Index Admiral & Vanguard Index & 25.75\% & 13.03\% & 14.19\% \\
Vanguard Small-Cap Index Admiral & Vanguard Index & 15.17\% & 7.25\% & 8.44\% \\
Vanguard Total Stock Market Index Admiral & Vanguard Index & 18.25\% & 11.25\% & 12.29\% \\
Vanguard Value Index Admiral & Vanguard Index & 10.19\% & 9.79\% & 10.90\% \\
Vanguard Wellington Investor & J. Keogh, T. Martin & 10.27\% & 7.60\% & 8.60\% \\
Vanguard Windsor II Investor & J. Keogh, T. Martin & 9.64\% & 8.53\% & 9.84\% \\
Vanguard Windsor Investor & J. Keogh, T. Martin & 8.67\% & 8.42\% & 9.52\% \\
Vanguard Wellesley Income Investor & J. Keogh, T. Martin & 6.31\% & 5.32\% & 6.19\% \\
Vanguard World Stock Index Investor & Vanguard Index & 20.64\% & 12.12\% & 12.83\% \\

\bottomrule
\end{longtable}

\twocolumn

\end{document}